\documentclass{bmvc2k}

\usepackage{amsmath,amssymb} 
\usepackage{todonotes}
\usepackage{tabularx}
\usepackage{xfrac}

\title{Pseudo-Label Noise Suppression Techniques for Semi-Supervised Semantic Segmentation}

\addauthor{Sebastian Scherer}{sebastian.scherer1@uni-a.de}{1}
\addauthor{Robin Schön}{robin.schoen@uni-a.de}{1}
\addauthor{Rainer Lienhart}{rainer.lienhart@uni-a.de}{1}

\addinstitution{
 University of Augsburg\\
 Machine Learning and Computer Vision Lab\\
 Augsburg, Germany
}

\runninghead{Scherer, Schön, Lienhart}{Semi-Supervised Semantic Segmentation}

\def\etal{\emph{et al}\bmvaOneDot}

\begin{document}

\maketitle

\begin{abstract}
Semi-supervised learning (SSL) can reduce the need for large labelled datasets by incorporating unlabelled data into the training. This is particularly interesting for semantic segmentation, where labelling data is very costly and time-consuming. 
Current SSL approaches use an initially supervised trained model to generate predictions for unlabelled images, called pseudo-labels, which are subsequently used for training a new model from scratch. Since the predictions usually do not come from an error-free neural network, they are naturally full of errors. However, training with partially incorrect labels often reduce the final model performance. Thus, it is crucial to manage errors/noise of pseudo-labels wisely. In this work, we use three mechanisms to control pseudo-label noise and errors: 
(1) We construct a solid base framework by mixing images with cow-patterns on unlabelled images to reduce the negative impact of wrong pseudo-labels. Nevertheless, wrong pseudo-labels still have a negative impact on the performance. Therefore, (2) we propose a simple and effective loss weighting scheme for pseudo-labels defined by the feedback of the model trained on these pseudo-labels. This allows us to soft-weight the pseudo-label training examples based on their determined confidence score during training. 
(3) We also study the common practice to ignore pseudo-labels with low confidence and empirically analyse the influence and effect of pseudo-labels with different confidence ranges on SSL and the contribution of pseudo-label filtering to the achievable performance gains. 
We show that our method performs superior to state of-the-art alternatives on various datasets. Furthermore, we show that our findings also transfer to other tasks such as human pose estimation.
Our code is available at \url{https://github.com/ChristmasFan/SSL_Denoising_Segmentation}.

\end{abstract}

\section{Introduction}
\label{sec:intro}
Semantic segmentation has accomplished amazing performance on annotated data and has become an important task in computer vision. 
However, labelling data for semantic segmentation requires assigning a class label to each pixel in an image, which is an extremely tedious and time-consuming task. As an example, annotating a single image of the Cityscapes dataset~\cite{cordts2016cityscapes}, which consists of images of urban scenes, takes up to 90 minutes~\cite{cordts2016cityscapes}. However, simply obtaining images is by far not as time-consuming. The goal of semi-supervised learning (SSL) is to incorporate unlabelled data into standard supervised training, improving overall robustness and performance while reducing the amount of labelled data needed. 

In SSL, most existing approaches treat model predictions as ground truth for unlabelled images during training, while human-generated annotations are used for labelled images. In this context, the predictions from a model that will be used as targets in the loss calculation are called \textit{pseudo-labels}. Due to the fact that the predictions and thus the pseudo-labels are inevitably error-prone, the network is partly trained on wrong annotations. Such pseudo-labels are sometimes called noisy labels, and DNNs are susceptible to them. There is usually a large gap between networks trained on clean and noisy datasets~\cite{song2020learning_noisy, cordeiro2020survey_noisy}.

In this paper, three mechanisms to handle error-prone pseudo-labels from unlabelled data will be analysed to improve the performance for semantic segmentation. Among them, we study the common pseudo-label filtering technique as well as a novel loss function that weights pseudo-labels according to the feedback of the trained model to reduce the influence of wrong pseudo-labels during training. This novel weighted loss function boosts SSL performance substantially and is a core contribution of this paper. We analyse this approach on several datasets to build a general and better understanding. We then integrate it into a naive pseudo-label refinement method and combine it with a cow-pattern based image mixing for data augmentation. Furthermore, we demonstrate the effectiveness of our approach on several datasets. Specially, we provide results at a very low data regime and show that we can significantly improve the performance when training e.g. only with 15 labelled examples on the Cityscapes dataset~\cite{cordts2016cityscapes}. We also demonstrate the effectiveness of our proposed mechanisms on a different task than semantic segmentation, namely human pose estimation.

\section{Related Work}
\normalfont{\textbf{Semi-Supervised Learning (SSL)}} aims to train a model using labelled and unlabelled images. Current SSL approaches utilise the prediction of a model as supervision for the training on unlabelled data and can be weakly separated into consistency regularization~\cite{sohn2020fixmatch, berthelot2019mixmatch, berthelot2019remixmatch, zhang2021flexmatch, french2019semi} and self-training approaches~\cite{yuan2021simple, Pseudo-Labels_Teacher-Student}. Both have the usage of pseudo-labels in common. Typically, the parameters of the model that generates pseudo-labels is fixed at a training iteration in self-training, while it changes at consistency regularization. Both approaches utilise strong data augmentation such as CutOut~\cite{devries2017improved}, CutMix~\cite{french2019semi}, ClassMix~\cite{olsson2021classmix} and colour space augmentations~\cite{sohn2020fixmatch, yuan2021simple, zou2020pseudoseg} when training with pseudo-labels. In this context, data augmentation is called perturbation because it makes it difficult for the model to reproduce the pseudo-label prediction. It is known that utilising perturbations has a strong effect on the final performance. 

Some methods further apply confidence-based filtering of pseudo labels~\cite{sohn2020fixmatch, zhang2021flexmatch, zhou2020uncertainty, zheng2021rectifying, yang2022st++, rizve2021defense}. FixMatch~\cite{sohn2020fixmatch} uses a fixed confidence threshold for all classes to remove uncertain labels for image classification. FlexMatch~\cite{zhang2021flexmatch} further extends this by filtering low-confidence samples with class-wise thresholds. Other approaches use different methods than model confidence to measure uncertainty. Zhou~\etal~\cite{zhou2020uncertainty} use multiple forward passes of different augmented version of the same image to estimate uncertainty and filter pseudo-labels. 
Yang~\etal~\cite{yang2022st++} uses multiple model training snapshots to determine a measure of uncertainty. 
However, none of them apply a comprehensive study on the effect of confidence-based filtering for semantic segmentation. 
\\
Our work is mostly related to those of~\cite{yuan2021simple}. In comparison to their work, we use simple image mixing with a diverse and dynamic mask instead of data augmentation strategies on pixel-level. We further apply pseudo-label filtering and dynamic weighting as denoising strategies to handle error-prone pseudo-labels. All approaches are examined in more detail on several datasets. 

\section{Methodology}
\label{sec:methodology}
In this section we first define import notation and the problem to solve in Sec.~\ref{subsec:overview}. Then we introduce the general SSL framework in Sec.~\ref{subsec:base_method}. Finally, we describe pseudo-labels filtering in Sec.~\ref{subsec:pseudo_label_filtering}. Our loss function with a dynamical weighting scheme is described at Sec.~\ref{subsec:robust_loss}.

\subsection{Problem Definition}
\label{subsec:overview}
Let $\mathcal{D}$ be a dataset consisting of a set $\mathcal{D}_\mathcal{L}$ of labelled and a set $\mathcal{D}_\mathcal{U}$ of unlabelled images. Their respective sizes are $N_l$ and $N_\mathcal{U}$. $\mathbf{x}_l \in \mathcal{D}_\mathcal{L}$ and $\mathbf{x}_u \in \mathcal{D}_\mathcal{U}$ denote any labelled and unlabelled image from the corresponding set of images. The labelled dataset has access to the ground truth segmentation masks $\mathbf{y}_l$, i.e., $\mathcal{D}_\mathcal{L} = \{ (\mathbf{x}_l^i, \mathbf{y}_l^i) \}^{N_l}_{i=1}$. The unlabelled dataset $\mathcal{D}_\mathcal{U}$ has no segmentation masks, i.e., $\mathcal{D}_\mathcal{U} = \{ (\mathbf{x}_u^i)\})^{N_u}_{i=1}$. To receive training feedback for unlabelled images, we use the predictions of a segmentation model as pseudo-labels $\hat{\mathbf{y}}$.  $\mathbf{y}$, in contrast, denotes ground truth annotations from the dataset. The aim is to improve the performance of the segmentation model by exploiting unlabelled images with generated pseudo-labels.

\subsection{SSL Base Approach}
\label{subsec:base_method}
We follow the simple idea of iterative pseudo-label refinement~\cite{Pseudo-Labels_Teacher-Student, NaiveStudent} and first train an initial segmentation model on the labelled dataset.
We then use the model as teacher $F_T$ to generate pseudo-labels for the unlabelled images by taking the argmax over the class dimension of the prediction $F_T(\mathbf{x}_u)$ for all input images $\mathbf{x}_u$, i.e., $\hat{\mathbf{y}}= argmax(F_T(\mathbf{x}_u))$. 
Then we train a student model $F_S$ with identical architecture as the teacher model $F_T$ on the labelled and unlabelled dataset $D = \{ (\mathbf{x}_l^i, \mathbf{y}_l^i) \}^{N_l}_{i=1} \cup \: \{ (\mathbf{x}_u^i, \mathbf{\hat{y}}_u^i) \}^{N_u}_{i=1}$ from scratch. After  training, the pseudo-label dataset is updated by the predictions of the student model and a new model is trained from scratch. This process can be repeated several times.

\begin{figure*}[t]
    \centering
    \begin{tabularx}{\linewidth}{*{4}{l@{ }}}
         \includegraphics[width=0.220\linewidth]{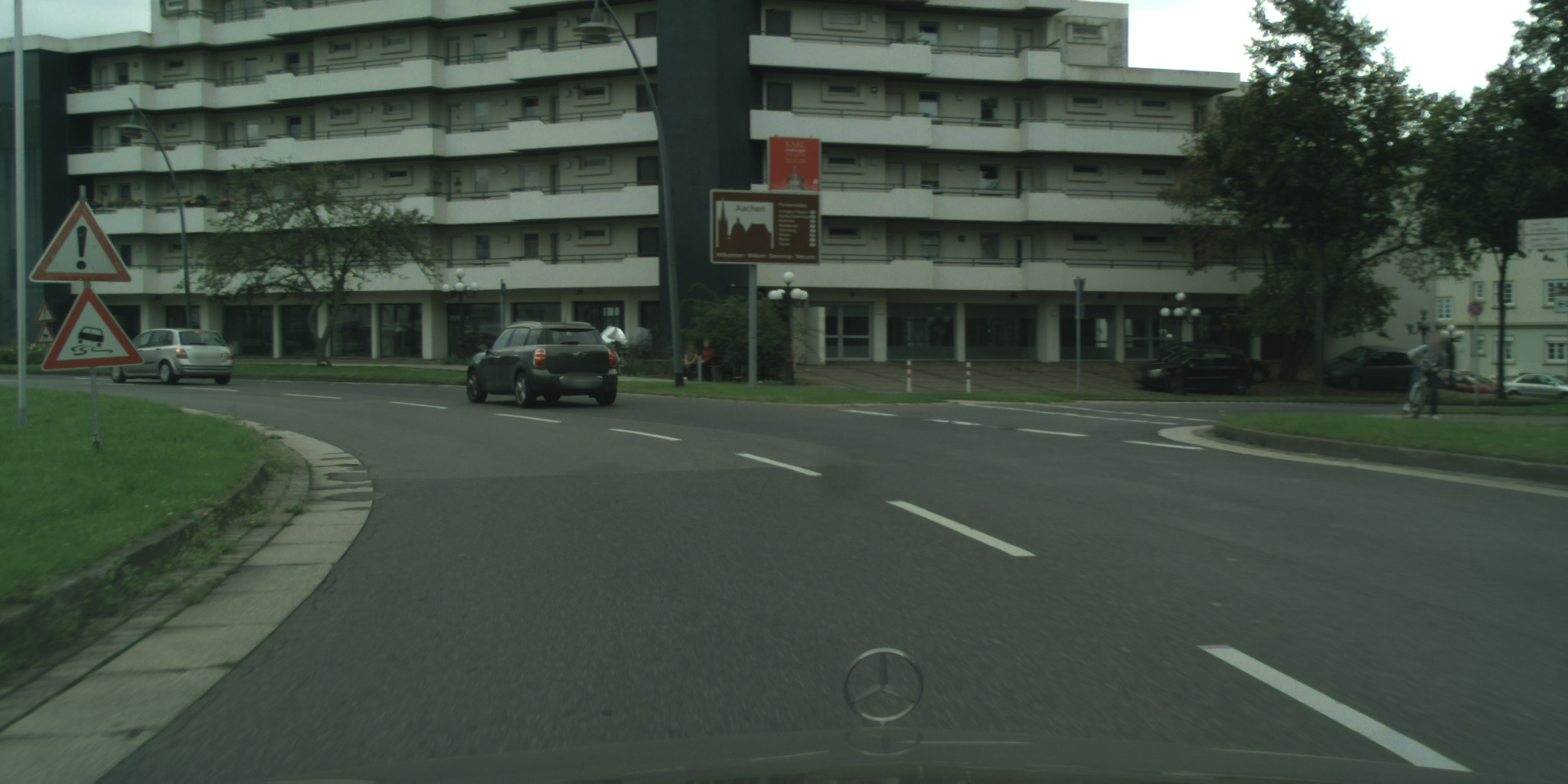} &
         \includegraphics[width=0.220\linewidth]{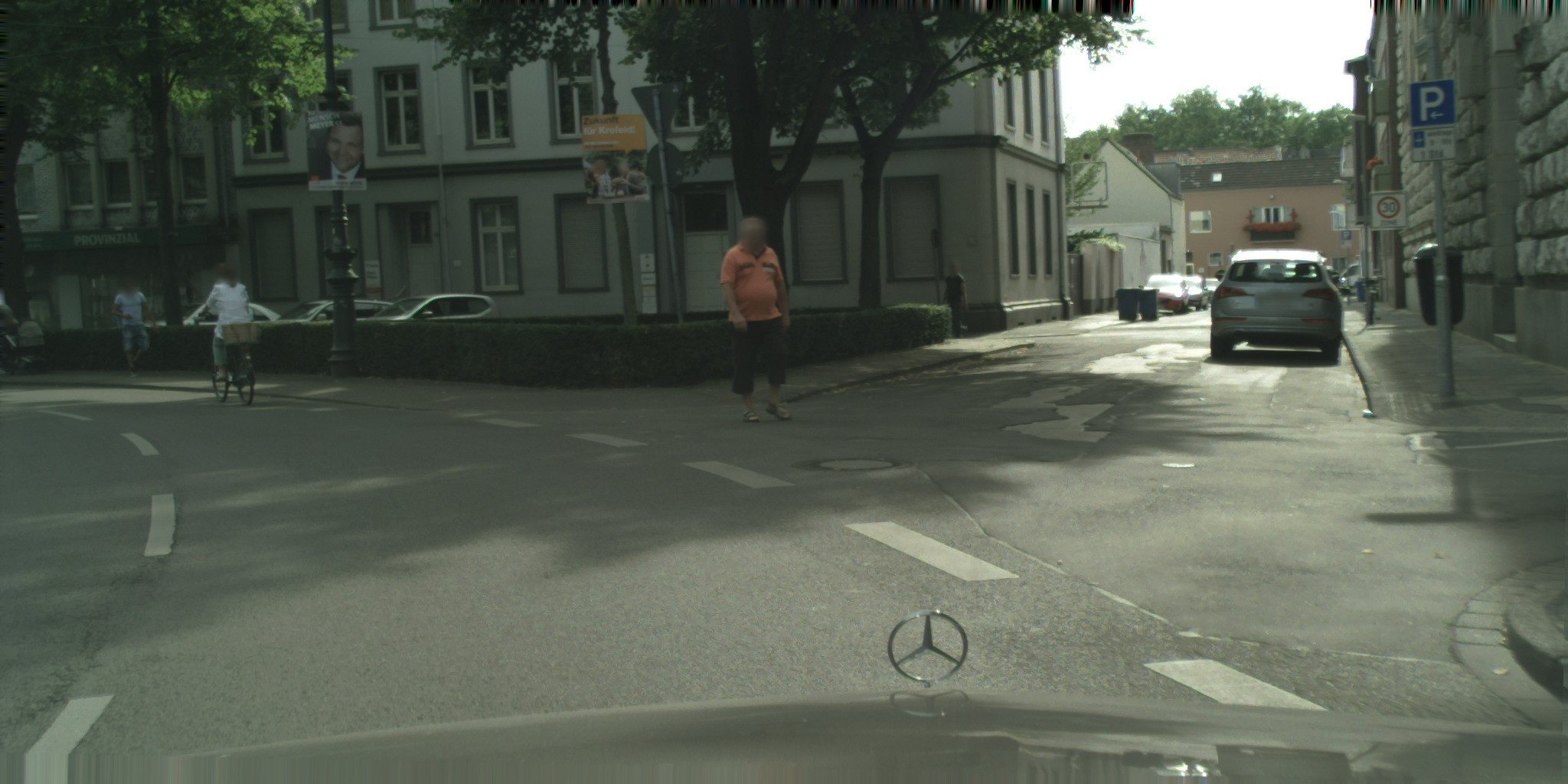} &
         \includegraphics[width=0.220\linewidth]{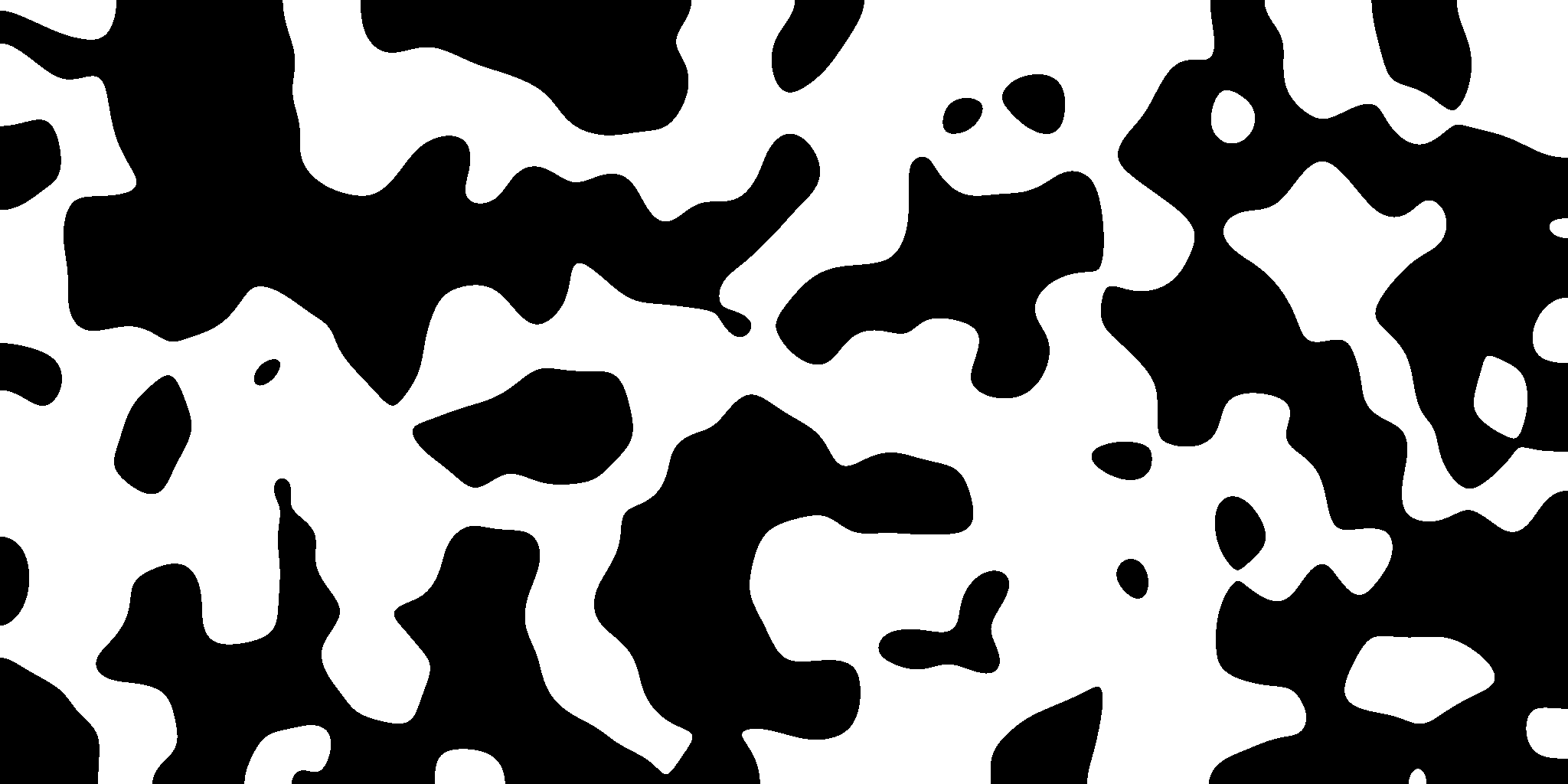} &
         \includegraphics[width=0.220\linewidth]{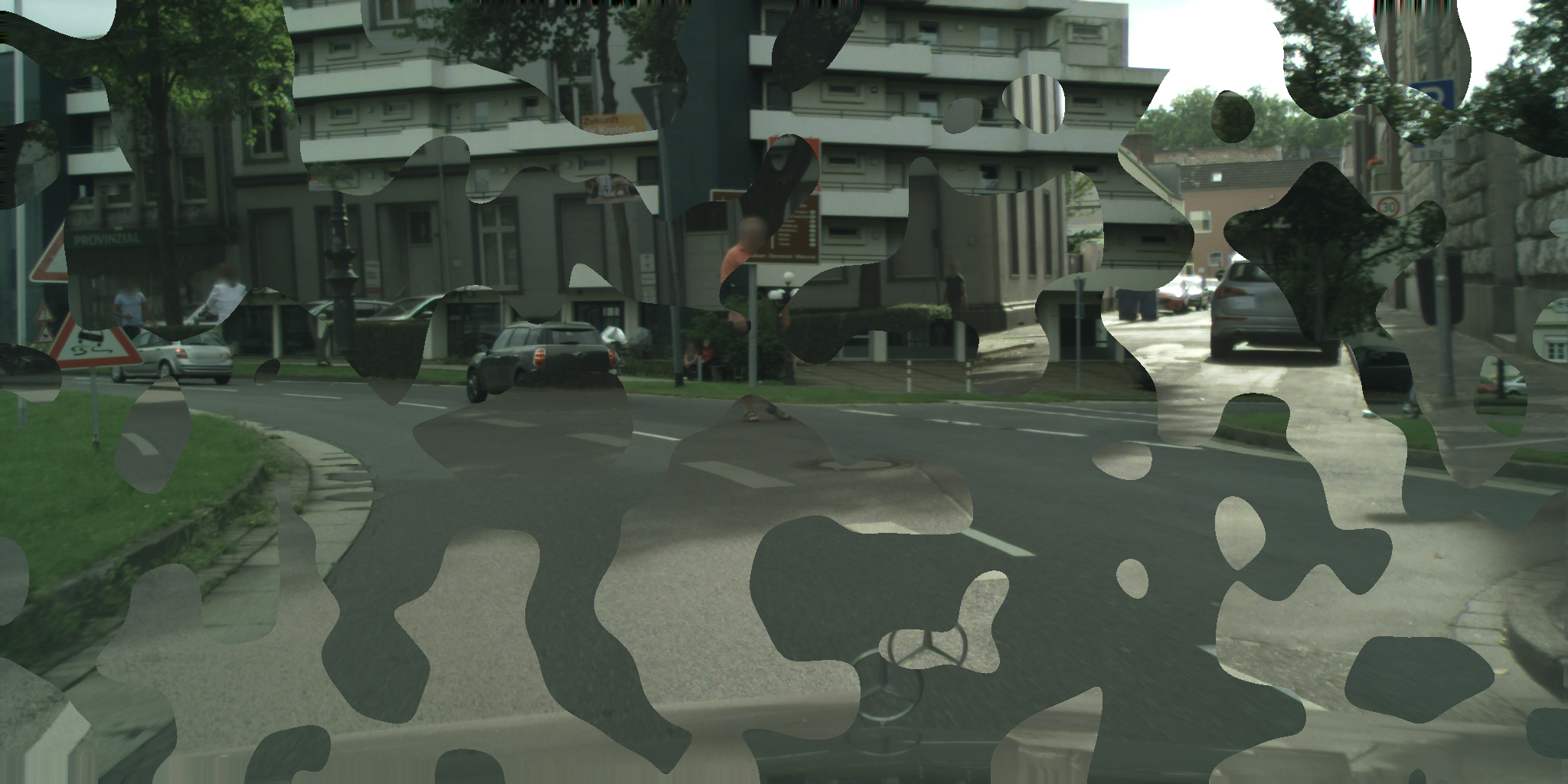} 
        \\
         (a) first image
        &(b) second image
        &(c) cow mask
        &(d) resulting image
    \end{tabularx}
    \caption{Example of the image mixing perturbation. The teacher network produces pseudo-labels on the original images. The pseudo-labels are mixed identical as the images. The student network is then trained to produce the mixed pseudo-labels on the mixed image.}
    \label{fig:cowmask}
\end{figure*}

Previous work based on consistency regularization has shown that adding perturbations at the image space when training with pseudo-labels helps to improve the performance~\cite{sohn2020fixmatch, french2019semi, olsson2021classmix}. Such perturbations are usually data augmentation strategies. They can prevent the model from easily memorizing the noisy pseudo-labels.  
Another motivation for the perturbations is that it may avoid ending up in the same local minima as the teacher during training.
Motivated by those works, we utilise image mixing as perturbation for semantic segmentation when training with pseudo-labels. 
Here the image is changed by the Cutmix~\cite{cutmixyun2019} augmentation using a CowMask~\cite{french2020milking}. With this augmentation, individual parts of the image are replaced by another image. The CowMask enables a greater diversity compared to CutMix~\cite{cutmixyun2019} while looking similar to the typical black and white skin pattern of a cow. To generate a training sample of unlabelled data, we take two unlabelled images $\mathbf{x}^1_u$ and $\mathbf{x}^2_u$ and calculate their pseudo-labels $\hat{\mathbf{y}}^1_u$ and $\hat{\mathbf{y}}^2_u$ with the teacher model.
Based on a binary CowMask $M \in \{0,1\} ^{W \times H}$, a new augmented image $\bar{\mathbf{x}}_u$ and its pseudo-label is calculated by:
\begin{equation}
\begin{split}
    \bar{\mathbf{x}}_u = M \odot \mathbf{x}_u^1 + (1-M) \odot \mathbf{x}_u^2, \\
    \hat{\mathbf{y}}_u = M \odot \hat{\mathbf{y}}_u^1 + (1-M) \odot \hat{\mathbf{y}}_u^1,
\end{split}
\end{equation}
where $\odot$ is the element-wise multiplication. An illustration for this image mixing method is shown in Figure~\ref{fig:cowmask}. For the calculation of a random CowMask, please refer to~\cite{french2020milking}. 
This perturbation simulates an occlusion of objects, as individual object parts are not visible within the image. 
It also simulates an additional segmentation task defined by the cow mask $M$.

\subsection{Pseudo-Label Filtering}
\label{subsec:pseudo_label_filtering}
As stated in~\cite{song2020learning_noisy}, wrong/noisy pseudo-labels in the training dataset generally have a negative impact on the training process, in the sense that the final performance is worse than if the false pseudo-labels were removed. This is due to that neural networks can overfit to wrong labels owing to their high capacity in totally memorising training samples~\cite{song2020learning_noisy, cordeiro2020survey_noisy}. We made the same observation in an iteration of our method on assumed unlabelled data using pseudo labels. If all wrong pseudo-labels were not used as feedback at training, the subsequent performance increases enormously, which shows that wrong pseudo-labels do indeed lead to a poorer performing model.
The critical question is: \textit{How can we identify wrong pseudo-labels?} The common idea is to remove pseudo labels with low confidence from the training and assume that a low confidence correlates with wrong pseudo-labels. The confidence is defined by the output probability after the softmax function of the predicted class of the model. 
A higher confidence is equated with a higher certainty that the pseudo-label is correct.
Another group of methods~\cite{zhou2020uncertainty, zheng2021rectifying, rizve2021defense} measures uncertainty based on several different predictions per image. In doing so, we assume that the more the different predictions deviate from each other, the higher the uncertainty is. The different predictions can be generated, for example, by gathering the predictions of a model using different augmented versions of the same image,
by  using different dropout configurations within the model while forwarding the same image several times or
by using other independently trained models where each model is trained with a different data-stream. 
Given a set of predictions in the form of probability distributions for the same original image, we can average the distributions and take their probability at the pseudo-label index as a measure of confidence. 
We have also found that pixels at the class boundaries of the segmentation mask in particular, usually object edges, have higher uncertainty within an image than distant pixels.
Therefore, handling all pixel-wise pseudo-labels identical may be misleading. 
Motivated by this, we propose to include the information of the segmentation borders of the image space at the pseudo-label filtering process. We do this by dividing the pixel-wise pseudo-labels into two groups, depending on the distance to the nearest segmentation boundary.

\subsection{Pseudo-Label Weighting}
\label{subsec:robust_loss}
Recent studies on the memorisation effect of deep neural networks show that they first memorise training data of clean labels and then those of noisy labels~\cite{cordeiro2020survey_noisy, han2018co}. This leads to our observation that the confidences of correct label predictions grow faster than the confidences of wrong label predictions over the course of model training. Motivated by this observation, we propose a weighting scheme with the aim to assign smaller weights to possibly wrong pseudo-labels and therefore reduce the contribution of noisy pseudo-labels at  training. 
We calculate the weight of a pseudo-label from the current confidence value of the student model for the given pseudo-label. 
Similar to previous work~\cite{yuan2021simple} we further adapt the Symmetric Cross Entropy loss~\cite{wang2019symmetric} (SCE), since it has been shown that it improve the performance when training on noisy datasets. The resulting loss $l^i \in \mathbb{R}^{H \times W}$ for an image of shape $H \times W$ is defined as:
\begin{equation}
    l^i = \mathbf{w}^i \odot (\alpha l_{ce}(\mathbf{p}^i, \hat{\mathbf{y}}_u^i) + \beta l_{ce}(\hat{\mathbf{y}}_u^i, \mathbf{p}^i))
\end{equation}
where $\alpha$ and $\beta$ are balancing coefficients and set to 2.0 and 1.0 respectively. $\mathbf{p}^i$ is the prediction of the student model on the perturbed image $\bar{\mathbf{x}}_u^i$, i.e. $\mathbf{p}^i = F_S(\bar{\mathbf{x}}_u^i)$. $\hat{\mathbf{y}}_u^i$ is the pseudo label from the current teacher model, i.e. $\hat{\mathbf{y}}_u^i = argmax(F_T(\mathbf{x}_u^i))$.
The dynamic weights $\mathbf{w^i} \in \mathbb{R}^{H \times W}$ is defined by the confidence of the softmax output of the student model $F_S$ and calculated as follows:
\begin{equation}
    \mathbf{w}^i = P(\hat{\mathbf{y}}_u^i) \quad  \textrm{with} \quad P=F_S(\mathbf{x}_u^i).
\end{equation}
Compared to pseudo-label filtering, the proposed weighting scheme does not have the disadvantage that the resulting pseudo-label dataset is smaller than before, since no labels are removed.
Furthermore, compared to the pseudo-label filtering approach, we use the confidence from the trained model as feedback instead of the confidence from the teacher model.

\section{Experiments}
\label{sec:experiments}
\subsection{Implementation Details}
\normalfont{\textbf{Dataset.}}
We evaluate our framework on three different datasets, Cityscapes~\cite{cordts2016cityscapes}, PASCAL VOC 2012~\cite{everingham2010pascal} and Mapillary~\cite{neuhold2017mapillary}. The Cityscapes dataset contains of images with 19 individual classes, where 2975 and 500 are used for training and validation respectively. 
We use an image size of $1024 \times 2048$ for our experiments. We further provide results for the PASCAL VOC 2012~\cite{everingham2010pascal} dataset, which consists of 1464 training and 1449 validation images. Following~\cite{yuan2021simple, liu2022perturbed, wang2022semi}, we use SBD~\cite{hariharan2011semantic} as the augmented set with 9118 additional training images. We follow common practise and use center-crops of size $513 \times 513$.
We further incorporate the Mapillary~\cite{neuhold2017mapillary} dataset for our ablation study, since it contains 18000 diverse training images and 2000 images for validation. We additionally chose Mapillary because it contains a larger amount of diverse images.
We resize the images to $512 \times 1024$ and incorporate the same 19 classes as the Cityscapes dataset. For each dataset, we randomly sample different partitions from the training set and set them as labelled images, while using the rest of the images as unlabelled. We also analyse our solution for the case that only very few labelled data are available, for example 15 labelled images at Cityscapes. Those images are taken such that all 19 classes are highly present.

\normalfont{\textbf{Training.}}
Following previous work~\cite{yuan2021simple, liu2022perturbed, wang2022semi}, we use the DeepLabv3+ model~\cite{chen2017deeplab} with a ResNet101~\cite{he2016deep} backbone pretrained on Imagenet~\cite{deng2009imagenet}, if not stated otherwise. For the Cityscapes experiments with 15 labelled images, we initialize the backbone with COCO weights. The network is trained using stochastic gradient descent. 
We use identical training details as in~~\cite{yuan2021simple, wang2022semi}. We use the poly scheduling to decay the learning rate during the training. The learning rates are set to 0.001 and 0.01 for PASCAL VOC 2012 and Cityscapes dataset respectively. For the Cityscapes dataset, we use a crop-size of $769 \times 769$ during training. Batch size has been set to 8, where 2 samples are laballed and 6 unlabelled. We perform up to three iterations where we reinitialise the student network. 
We adopt the mean intersection over union (mIoU) as evaluation metric. Following ~\cite{yuan2021simple}, we apply random rescale and horizontal flip as augmentation. When using pseudo-labels for supervision, we additionally use the image mixing method.
To filter pseudo-labels according to their confidence, we first calculate a confidence histogram for each class from the pseudo-label dataset. From that we can calculate class-wise threshold to remove certain percentage of pseudo-labels. 

\subsection{Ablation Study}
To better investigate the influence of the individual methods we perform our ablation study on all three mentioned datasets.
We also specifically assume a scenario with only a few labelled samples, as this presents the greatest difficulty and has a very large difference in performance compared to the fully supervised model. There is also a higher number of wrong pseudo-labels, which makes pseudo-label filtering more interesting. 

\begin{table}[t]
\begin{center}
\begin{tabular}{| c | c c c c c c c || c | c | c |} 
 \hline
  Model & CM & ST  & ST$_{\scriptsize CM}$ &  PLF & PLW & PLW$_{\scriptsize SCE}$ &  Iter. & C\scriptsize(15) & M\scriptsize(100)  & P\scriptsize(183)  \\ [0.5ex] 
 \hline\hline
  DL3+ & &           &           &           &               &            & &           53.0 & 50.7 & 57.0   \\
  DL3+ &  \checkmark  &           &           & &             &           & & 52.1 & 50.2  & 56.2  \\
  DL3+ &  &           \checkmark  &           & &             &           & & 56.2 & 53.9 & 66.3  \\
  DL3+ &  &           &           \checkmark  & &             &           & & 60.3 & 56.1 & 68.1  \\
  DL3+ &  &           &           \checkmark  & \checkmark    &           & & & 60.9  & 56.8 & 67.8  \\
  DL3+ &  &           &           \checkmark  & &   \checkmark          &  & & 61.6  & 57.0 & 68.9\\
  DL3+ &  &           &           \checkmark  & &             & \checkmark  & & 62.4 & 57.8 & 69.5  \\
  DL3+ &  &           &           \checkmark  & \checkmark    &           & \checkmark & & 63.0 & 58.0 & 69.3 \\
  DL3+ &  &           &           \checkmark  & \checkmark    &           & \checkmark & \checkmark & 66.5 & 60.1 & -   \\

\hline 

\end{tabular}
\end{center}
\caption{ Ablation study on the proposed SSL framework. "C", "M", "P" stands for the Cityscapes, Mapillary and PASCAL  dataset respectively. The number in brackets stands for the number of labelled images used. 'CM' denotes CowMask augmentation. 'ST' denotes self-training with pseudo-labels. '$ST_{\scriptsize CM}$' uses image mixing augmentation for training on pseudo-labels. 'PLF' and 'PLW' refers to pseudo-label filtering and weighting. 'SCE' refers to the symmetric cross-entropy loss. }
\label{table:ablation}
\end{table}

\subsubsection{Impact of Image Mixing Augmentation}
In this section, we evaluate the contribution of the image mixing method. Table~\ref{table:ablation} shows on the one hand the resulting performance after a single iteration with  additional training on unlabelled data with pseudo-labels and on the other hand the performance when we further apply the image mixing method when training on pseudo-labels. 
At the Cityscapes dataset, using pseudo-labels improves the performance from $53.0 \%$ to $56.2 \%$. Further applying the image mixing approach, the performance increases to $60.2 \%$.
From this it can be seen that the use of an image mixing perturbation in addition to pseudo-labels leads to a similar performance boost as the use of pseudo-labels in general. To verify that the improvement does not come from the data augmentation strategy, we also trained the fully supervised baseline model with the image mixing augmentation. Here we can observe a small deterioration in performance, showing that this augmentation strategy only improves the result when applied to images with pseudo-labels, not to images where ground-truth annotation is available. The same findings can be seen at the PASCAL VOC 2012 and Mapillary dataset. 
Table ~\ref{table:CowMaskCutMix} further compares the usage of a CowMask to CutMix~\cite{cutmixyun2019}. It can be seen, that CowMask is more effective and yields better results.

\begin{figure*}
    \centering
    \begin{tabularx}{\linewidth}{*{2}{l@{ }}}
         \includegraphics[width=0.45\linewidth]{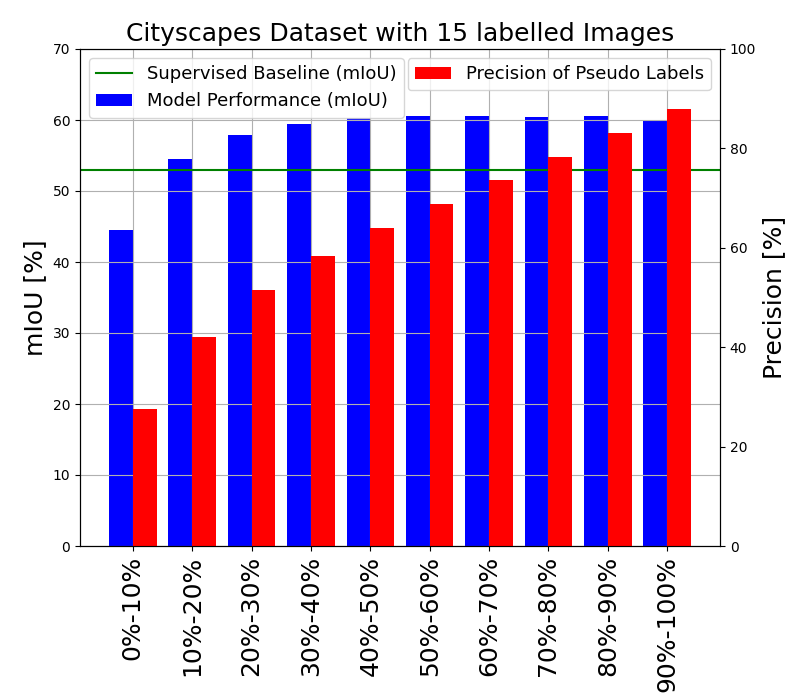} &
         \includegraphics[width=0.45\linewidth]{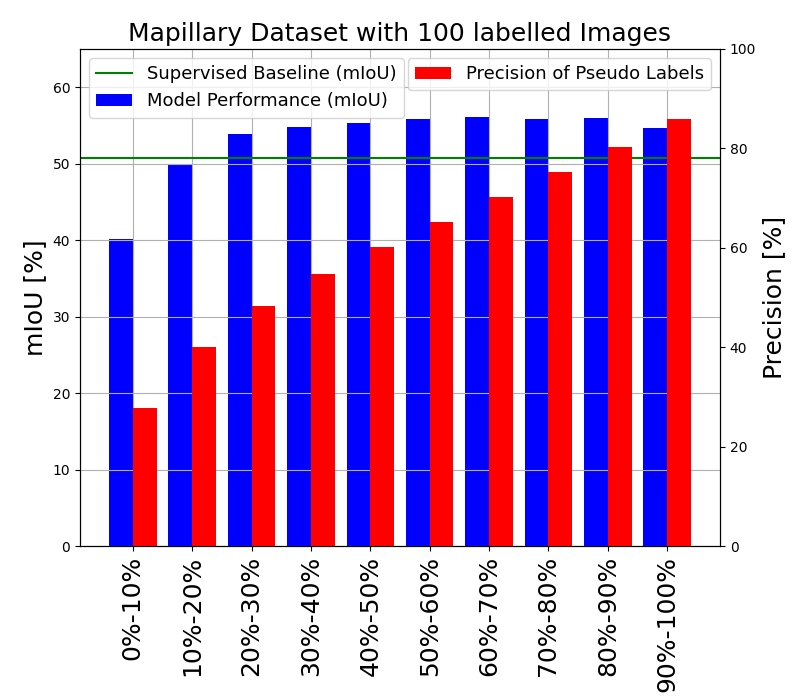} 
    \end{tabularx}
    \caption{Analysis of different pseudo-label datasets ordered in $10 \%$ steps according to confidence. We report the class-wise precision of those pseudo-labels as well as the model performance when incorporating a single pseudo-label subset into our SSL training. The green horizontal line shows the supervised only model.}
    \label{fig:exp_10}
\end{figure*}

\subsubsection{Impact of Pseudo-Label Filtering}
First we analysed whether the pseudo-label filtering by the softmax confidence of the model performs different as filtering based on several predictions as introduced in section~\ref{subsec:pseudo_label_filtering}. We did so by adding the most confident examples to a dataset and analyse this dataset for correctness. Here we observed that with all methods we obtain a dataset with equal quality. 
The multiple model approach also did not lead to an improvement, which probably comes from the fact that only a small amount of labelled data was available. Due to the small differences between the methods and the simplicity of the approach using the model confidence, we keep it for the rest of this work.

Next, we analysed pseudo-labels with different uncertainty in more detail. 
Given the complete pseudo-label dataset, we split it into ten subsets according to the confidence, taking into account different classes. 
For example, one dataset contains pseudo-labels of each class with the lowest confidence, i.e. $0 \% - 10 \%$.
Then we performed a semi-supervised training iteration using one of the ten pseudo-label datasets. 
Figure~\ref{fig:exp_10} shows the quality of the pseudo-label subsets and the resulting performance of the trained model when additionally using a single pseudo-label subset for training. As expected, we can observe that the confidence of the model prediction indeed correlates with the correctness of the pseudo-labels. For the Mapillary dataset, from the top $10 \%$ confident samples per class $86 \%$ are correct on average, while only $28 \%$ of the worst $10 \%$ are correct. A similar correlation can be seen at the Cityscapes experiment. 
Comparing the final performance when incorporating the pseudo-label subset into a semi-supervised training, we can observe a weaker correlation. For both datasets, taking the pseudo-labels with the worst $10 \%$ confidence decreases the final performance. 
At the Mapillary dataset we drop in performance from $50.7 \%$ to $\sim 40\%$ when we additionally use those pseudo-labels for training.
This shows that those pseudo-labels have a harmful influence at the training. The same can be observed at the Cityscapes experiment. 
However, this effect no longer occurs in samples with a slightly higher confidence, most probably because strong perturbations already suppress the negative effect of wrong pseudo-labels. 
At the Cityscapes experiment, using the weakest $10\% - 20\%$ samples additionally for training already improves the performance slightly. 
Even more interestingly, we can observe that the worst $30\% - 40\%$ have an almost equal effect in performance boost as the top $10\%$ confident samples ($90\%-100\%$) on both datasets. This is probably because the correct pseudo-labels from the lower confident subsets are samples that the model struggles with. In other words, the correct pseudo-labels from lower confident subsets improve the performance more than those from the high confident samples. High confident samples are probably more similar to examples from the labelled training set and correct pseudo-labels with low confidence have a higher impact on the resulting decision boundary. 

Based on this findings, we removed the most uncertain $20\%$ of each class from the complete pseudo-label dataset. The result is shown in Table~\ref{table:ablation}. The performance increases slightly for the Cityscapes and Mapillary experiment, respectively. 
At the PASCAL VOC experiment we can observe a small deterioration.
Here we also observed that additionally using the worst $10\%$ of the pseudo-labels lead to a similar performance as the supervised baseline.
From the experiments, it can be concluded that a confidence-based filtering does not lead to the hoped-for significant improvement. The Mapillary experiment further verified, that the marginal improvement does not come from the smaller dataset.
From these results, we conclude that removing incorrect pseudo-labels lead to an improvement, but removing correct pseudo-labels leads to a deterioration, and that both effects ultimately overlap. 
Thus, confidence-based filtering is not good enough at detecting wrong pseudo-labels because there are too many correct pseudo-labels with low confidence.

\begin{table}[t]
\begin{center}
\begin{tabular}{||c c c c c c c c ||} 
 \hline
  Method & Model & $\tfrac{1}{200}$ \scriptsize(15) &  $\tfrac{1}{20}$ \scriptsize(148) & $\tfrac{1}{16}$ \scriptsize(186)  & $\tfrac{1}{8}$ \scriptsize(372) & $\tfrac{1}{4}$ \scriptsize(744) &  Full \\ [0.5ex] 
 \hline\hline

    \hline
    SupOnly (ours) & DL3+ &   53.0     & 64.1 & 64.8   & 69.0   & 73.3 &   78.7 \\ 
    CutMix$^\star$~\cite{french2019semi} & DL3+ & - &  - & 67.1 & 71.8 & 76.4 & -  \\
    Yuan et al.~\cite{yuan2021simple} & DL3+   & - & - & - & 74.1 & 77.8 & 78.7 \\
    U2PL~\cite{wang2022semi} & DL3+  & -  & - & 74.9 & 76.5 & 78.5 & - \\
    PS-MT~\cite{liu2022perturbed} & DL3+  & -  & - & - & 77.1 & 78.4 & - \\
    CPS~\cite{CPS} & DL3+  & -  & - & 74.7 & 77.6 & \textbf{79.1} & 80.4 \\

     \hline
    Ours & DL3+ &  \textbf{66.5} &    \textbf{74.4} & \textbf{75.7} &   \textbf{78.0} & 78.7 &  -  \\
\hline 

\end{tabular}
\end{center}
\caption{Comparison of state-of-the-art methods for different portions of labelled data on the Cityscapes validation set. $^\star$ means approach has been reproduced by~\cite{wang2022semi}.}
\label{table:resuts}

\bigskip

\begin{center}
\begin{tabular}{||c c c c c c c ||} 
 \hline
  Method & Model  & $\tfrac{1}{16}$ \scriptsize(92) &  $\tfrac{1}{8}$ \scriptsize(183)  & $\tfrac{1}{4}$ \scriptsize(366)  & $\tfrac{1}{2}$ \scriptsize(732)  &  Full \scriptsize(1464) \\ [0.5ex] 
 \hline\hline

    SupOnly (ours) & DL3+ & 51.0  &  57.0  &  66.8   & 72.1  &  73.4  \\
    CutMix$^\star$~\cite{french2019semi} & DL3+ & 52.2 & 63.5    &  69.5    & 73.7 &  76.5   \\
    PseudoSeg~\cite{zou2020pseudoseg} & DL3+ & 57.6 & 65.5    &  69.1    & 72.4 &  73.2   \\
    PC$^2$Seg~\cite{zhong2021pixel} & DL3+ & 57.0 & 66.3      &  69.8    & 73.1 &  74.2   \\
    CPS~\cite{CPS} & DL3+ & 64.1 & 67.4      &  71.7    & 75.9 & -   \\
    U2PL~\cite{wang2022semi} & DL3+ & 68.0 & 69.2      &  73.7    & 76.4 & \textbf{79.5}   \\
    ST++~\cite{yang2022st++} & DL3+ & 65.2 & 71.0      &  \textbf{74.6}    & 77.3 & 79.1   \\
    \hline
    Ours & DL3+ & \textbf{68.6} &   \textbf{71.4}   & 74.4    & \textbf{77.5} &  \textbf{79.5}  \\
\hline 

\end{tabular}
\end{center}
\caption{Comparison of state-of-the-art methods for different portions of labelled data on the PASCAL VOC 2012 validation set. The labelled images are selected from the original VOC training dataset. The rest plus the images from the SBD~\cite{hariharan2011semantic} dataset are regarded as unlabelled data. $^\star$ means approach has been reproduced by~\cite{wang2022semi}.  }
\label{table:resuts_voc}
\end{table}

\subsubsection{Impact of the Robust Loss Function}
To improve the performance of the model and reduce the impact of wrong pseudo-labels, we employ a weighting scheme of pseudo-labels defined by the confidence of the trained model for the pseudo-labels of the teacher. 
From Table~\ref{table:ablation} we can see that the weighting scheme improves the performance on all three experimental settings.
We further verified whether wrong pseudo-labels indeed have a lower weight during training than correct pseudo-labels.
Here we found that the weights of the correct pseudo-labels were on average twice as high as those of wrong pseudo-labels. This shows that the influence of wrong pseudo-labels during training is reduced and therefore an improvement in performance occurs. Applying the SCE loss instead of the standard cross-entropy loss further improves the performance slightly. We provide sensitivity analysis of the alpha and beta parameters of the SCE loss in the Appendix.

\subsection{Comparison with State-of-the-art Methods}
We compare our method with the most recent semi-supervised semantic segmentation methods that are mostly related to our work: Simple Baseline~\cite{yuan2021simple}, U2PL~\cite{wang2022semi}, ST++~\cite{yang2022st++} and PS-MT~\cite{liu2022perturbed}. We further compare our solution to CPS~\cite{CPS} which is slightly different to our work, since here two independent networks are trained simultaneously. We report results without using pseudo-label filtering with the exception of 15 labelled image at Cityscapes.
The results for the Cityscapes dataset are shown in Table~\ref{table:resuts}. Our approach improves the performance from our supervised baseline by $+13.5 \%$, $+10.3 \%$, $+10.9 \%$, $+9 \%$ and $+5.4 \%$ under $\sfrac{1}{200}$, $\sfrac{1}{20}$, $\sfrac{1}{16}$, $\sfrac{1}{8}$ and $\sfrac{1}{4}$ partition protocols respectively. The proposed approach performs favourably against state-of-the art approaches that are mostly related to our work when assuming $\frac{1}{16}$, $\frac{1}{8}$ and $\frac{1}{4}$ to be labelled. We can notice that using $\frac{1}{4}$ of the labelled data in our SSL framework achieves a similar performance as using the whole dataset for supervision. 
Comparing the results to those of~\cite{yuan2021simple}, the improvement comes from the image mixing method using a CowMask and the pseudo-label weighting strategy, as shown in our ablation study. 
We further provide results on the classic PASCAL VOC 2012 dataset in Table~\ref{table:resuts_voc}. Here our approach improves the performance from our supervised baseline by $+17.6 \%$, $+14.4 \%$, $+7.6 \%$ and $+5.4 \%$ under $\sfrac{1}{16}$, $\sfrac{1}{8}$, $\sfrac{1}{4}$ and $\sfrac{1}{2}$ partition protocols respectively.

 \subsection{Human Pose Estimation}
To verify if our findings are not specific to semantic segmentation, we also apply our method on the human pose estimation (HPE) task. The experiments are conducted on the LSP dataset~\cite{Johnson10}. It contains of 2000 images with 14 annotated joints. For our ablations we use 1600 images as training set and the rest for evaluation. We use \textit{SimpleBaseline}~\cite{xiao2018simple} to estimate heatmaps and ResNet18 as its backbone for simplicity. Note that our approach can be applied to other pose estimators as well. 
We use the standard Percentage of Correct Keypoints (PCK) measurement that quantifies the fraction of correct predictions within an error threshold $\lambda$ normalised by the size of the torso.  We set $\lambda$ to 0.2, i.e. PCK@0.2.
We perform ablations with image mixing using a CowMask, pseudo label weighting as well as filtering. The results are summarised at Table~\ref{table:HPE}. Similar to semantic segmentation, we can observe an improvement when using image mixing as well as our proposed pseudo-label weighting scheme, while pseudo-label filtering is less effective. Since the findings for HPE are similar to those of semantic segmentation, we can assume that they also apply to other domains and tasks. 

\begin{figure*}
    \centering
    \begin{tabularx}{\linewidth}{*{5}{l@{ }}}
         \includegraphics[width=0.190\linewidth]{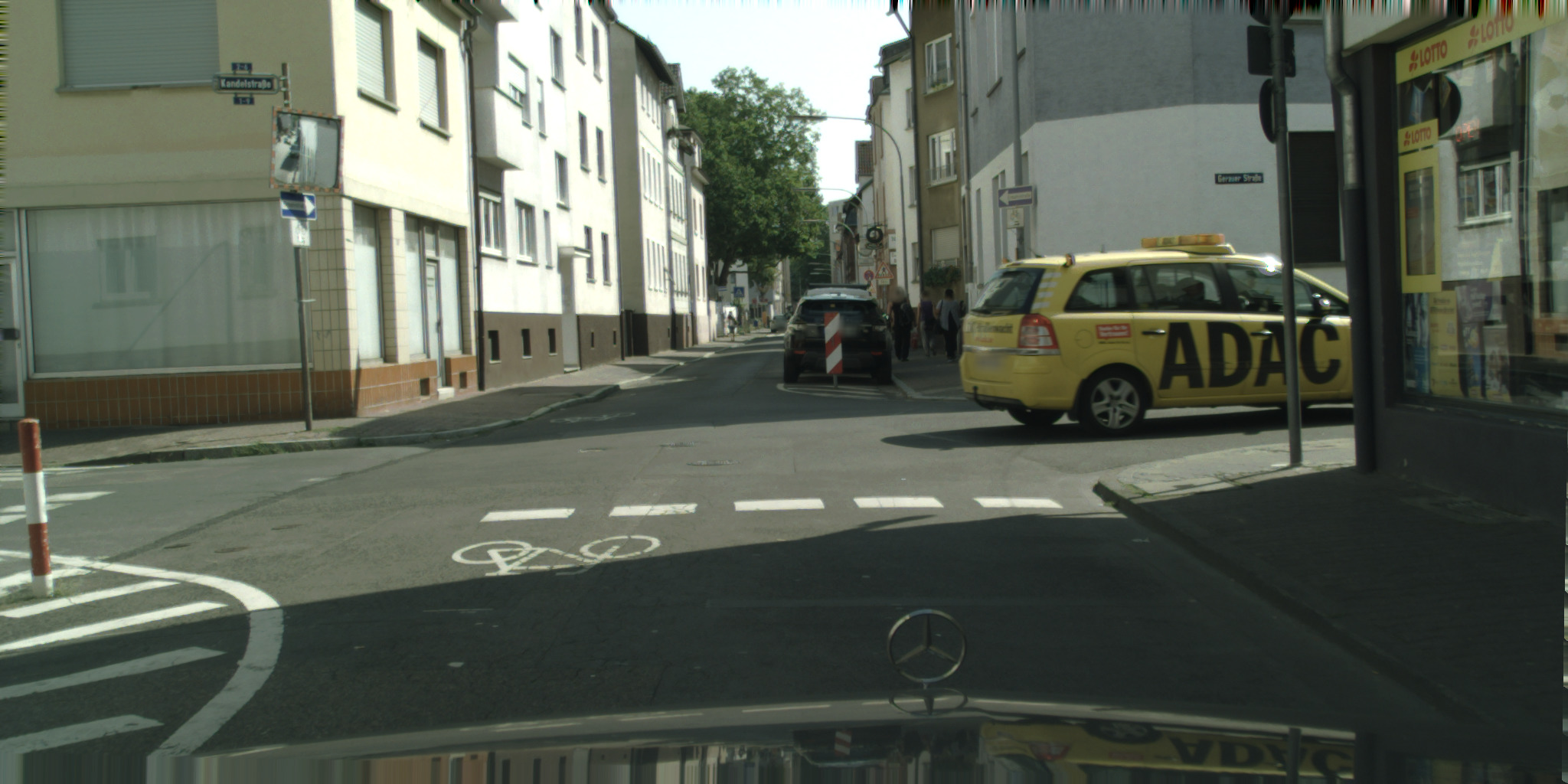} &
         \includegraphics[width=0.190\linewidth]{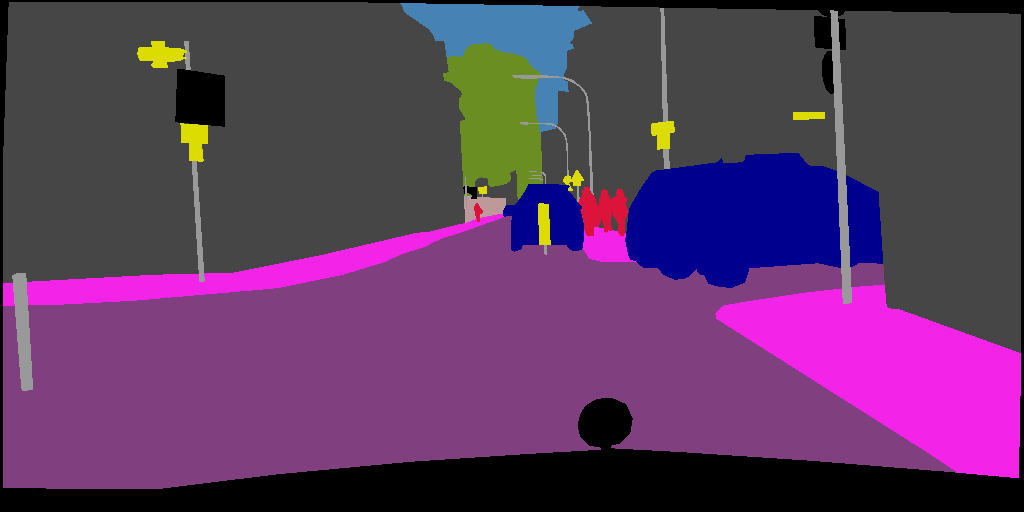} &
         \includegraphics[width=0.190\linewidth]{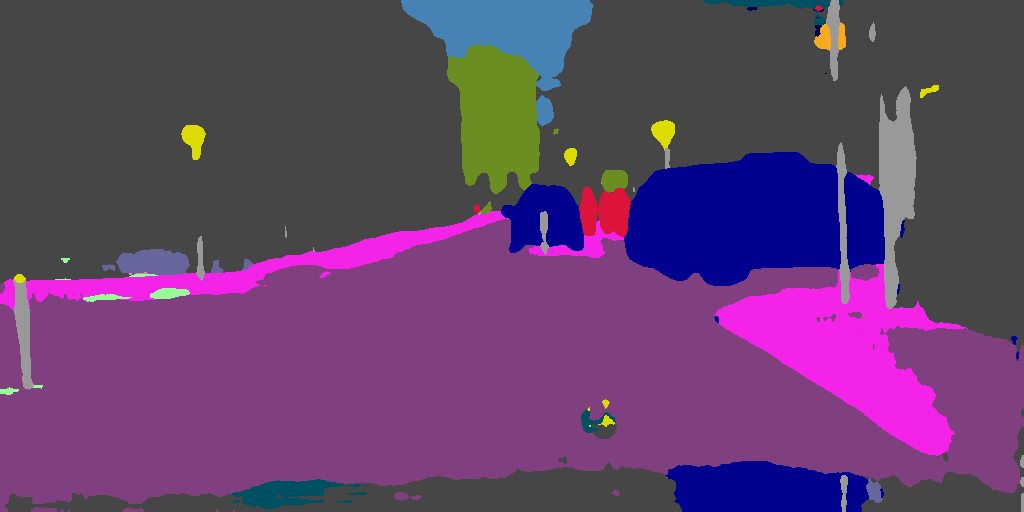} &
         \includegraphics[width=0.190\linewidth]{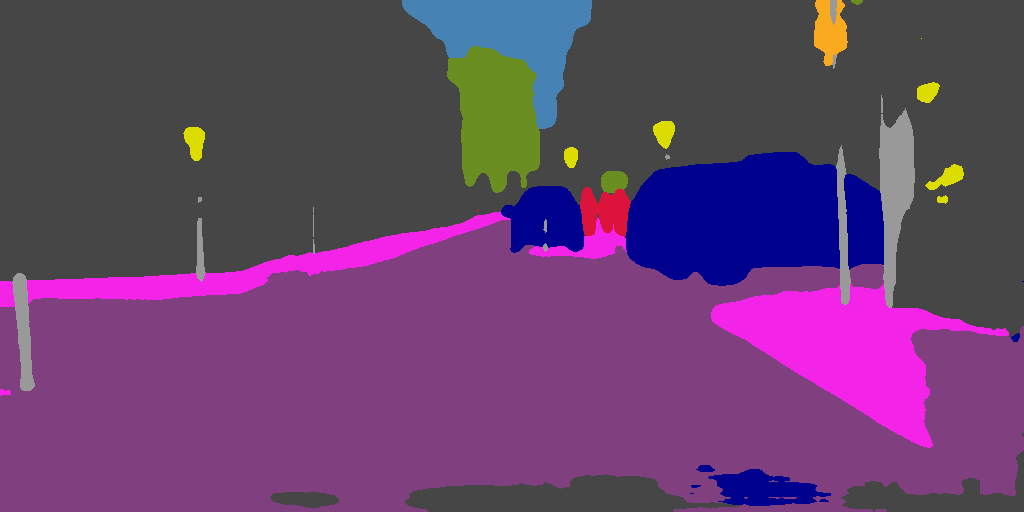} &
         \includegraphics[width=0.190\linewidth]{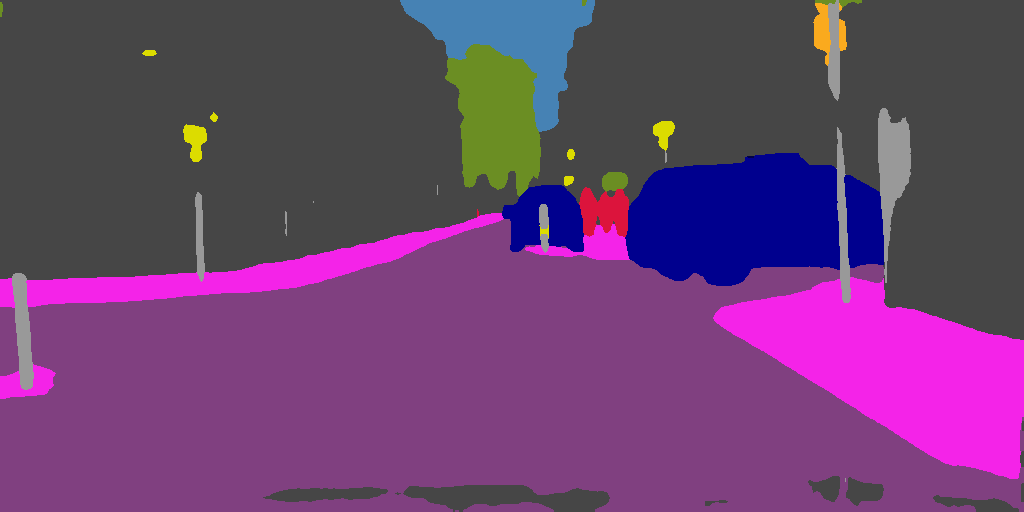} 
          \\
         \includegraphics[width=0.190\linewidth]{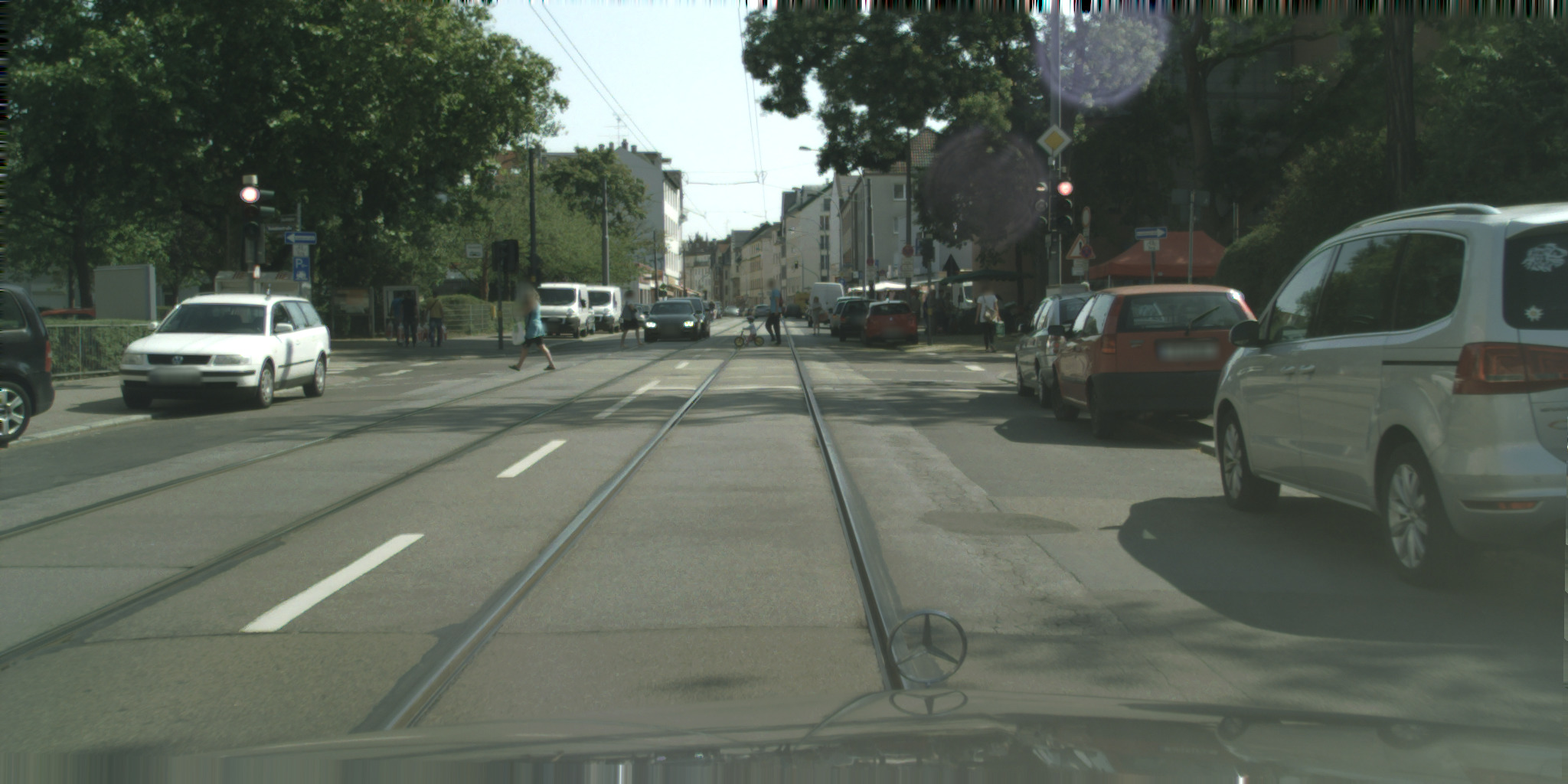} &
         \includegraphics[width=0.190\linewidth]{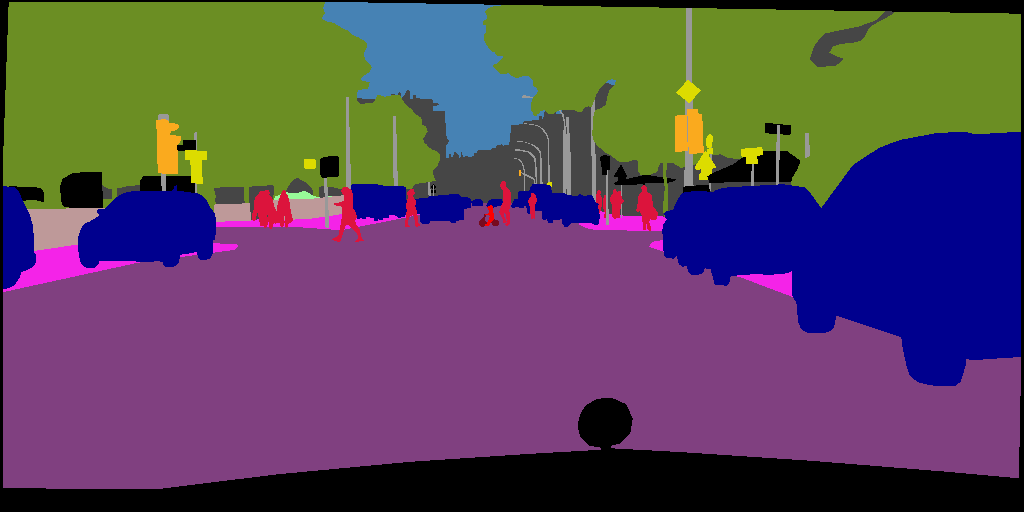} &
         \includegraphics[width=0.190\linewidth]{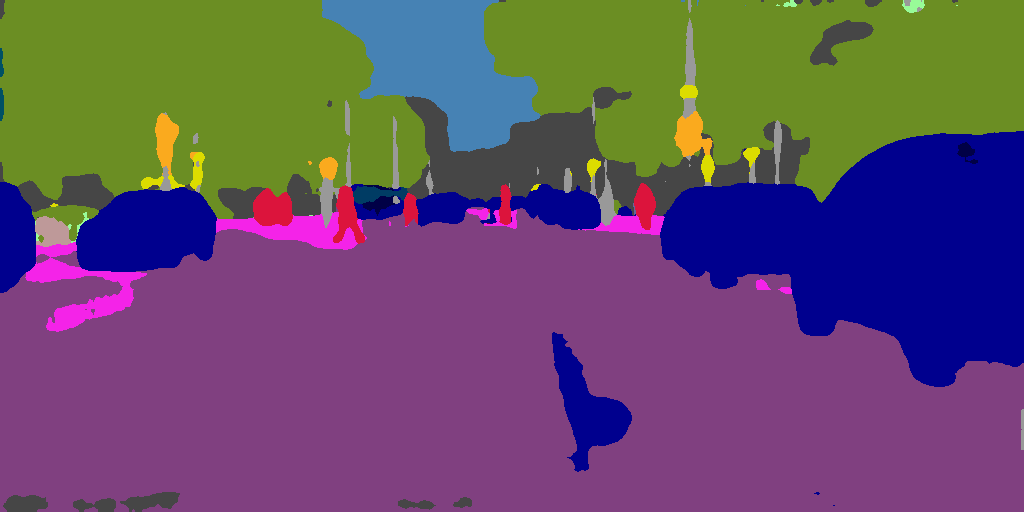} &
         \includegraphics[width=0.190\linewidth]{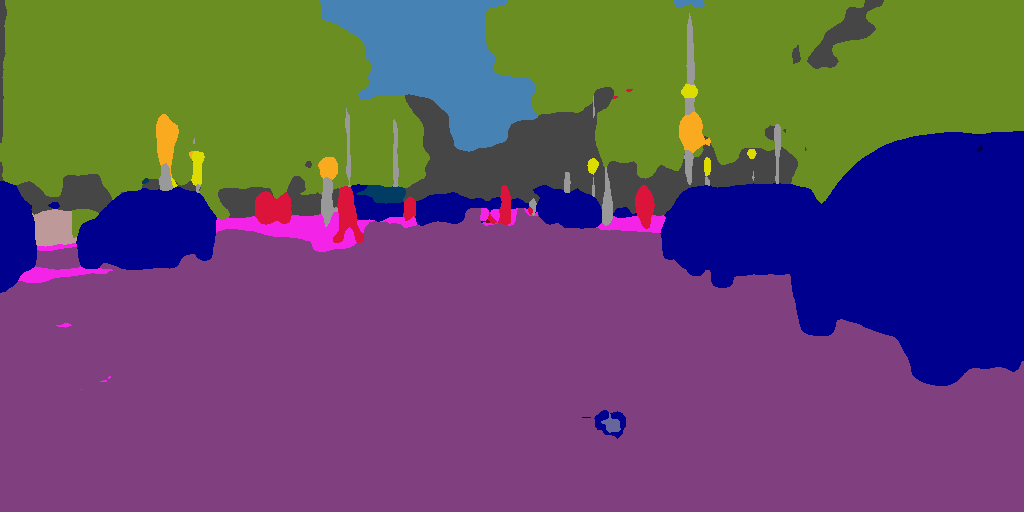} &
         \includegraphics[width=0.190\linewidth]{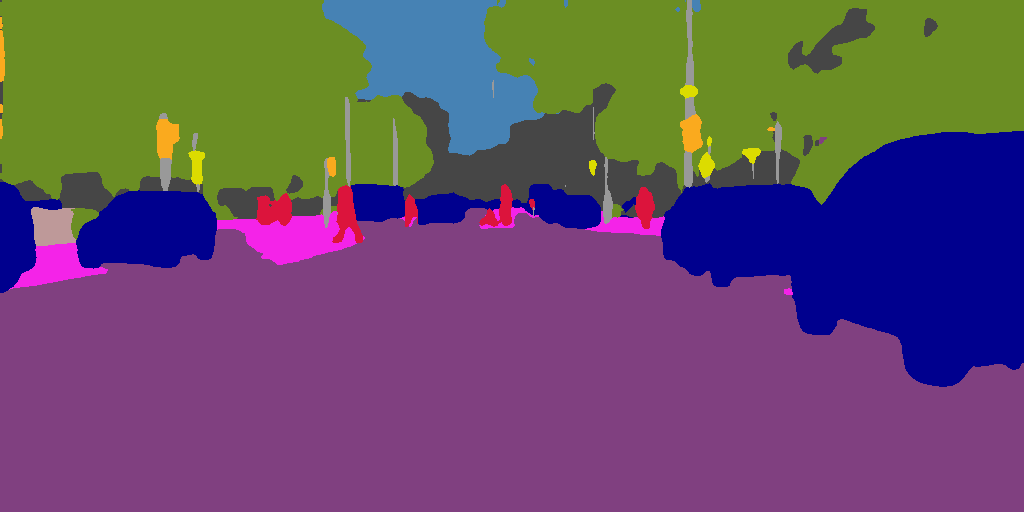} 
        \\
         (a) Image
        &(b) GT
        &(c) Baseline
        &(d) ST
        &(e) Ours
    \end{tabularx}
    \caption{Qualitative examples on the Cityscapes val set. All models are trained using the annotation of 15 images. The Baseline is trained supervised only. "ST" represents the naive self-training approach without our proposed extensions. Our proposed solution produces improved results compared to the  baseline and the naive self-training approach.}
    \label{fig:examples}
\end{figure*}

\begin{table}
    \parbox{.46\linewidth}{
    \centering
    \begin{tabular}{c|ccc}
    \hline
    \small Method & \small C$_{\scriptsize(15)}$ & \small M$_{\scriptsize(100)}$ & \small P$_{\scriptsize(183)}$ \\
    \hline
    \small ST & \small 56.2  & \small 53.9 & \small 66.3  \\
    \small ST + CutMix & \small 58.3 & \small 55.0 & \small 66.9 \\
    \small ST + CowMask & \small 60.3 & \small 56.1 & \small 68.1\\
    \hline
    \end{tabular}
    \caption{Ablation study comparing CowMask and CutMix augmentation on pseudo-labels at the first iteration. "C", "M", "P" refer to the different datasets.  \label{table:CowMaskCutMix}}
    }
\hfill
    \parbox{.48\linewidth}{
    \centering
    \begin{tabular}{c|ccc}
    \hline
    \small Method & \small $\frac{1}{16}_{\scriptsize(100)}$ & \small $\frac{1}{4}_{\scriptsize(400)}$ &  \small Full$_{\scriptsize(1600)}$\\
    \hline
    \small Sup. only & \small 62.3 & \small 73.5 & \small 82.6 \\
    \hline
    \small ST & \small 65.9 & \small 76.4 & - \\
    \small + CowMask & \small 68.2 & \small 78.2 & - \\
    \small + PLW & \small 69.6 & \small 78.8 & - \\
    \small + PLF & \small 69.4 & \small 78.5 & - \\
    \small + iter & \small 71.3 & \small 79.4 & - \\
    \hline
    
    \end{tabular}
    \caption{Ablation study of our proposed methods on a common HPE task. \label{table:HPE}}
    }
\end{table}

\section{Conclusion}
We have proposed a simple and efficient semi-supervised semantic segmentation framework. It employs efficient image mixing with a diverse mask and pseudo-label denoising strategies. We have analysed the effect of confident-based pseudo label filtering in more detail and shown that it leads to little, if any, improvement. In addition to the pseudo-label filtering we have also analysed our novel weighting of pseudo-labels given the current feedback of the model. This approach has shown to be effective and yield to an overall improvement on three different datasets. 
Quantitative and qualitative comparisons have shown that the proposed method performs favourably against state-of-the art approaches and also apply to other tasks such as human pose estimation.

\bibliography{egbib}
\end{document}